\documentclass[runningheads]{llncs}
\usepackage[misc]{ifsym}
\usepackage[T1]{fontenc}
\usepackage{amsmath,amssymb}
\usepackage{booktabs}
\usepackage[noend]{algpseudocode}
\usepackage{algorithm}
\usepackage{amssymb}
\usepackage{multirow}
\usepackage{booktabs}
\usepackage{cleveref}
\usepackage{comment}
\usepackage{wrapfig}
\usepackage{xcolor}
\usepackage{graphicx,subcaption}
\usepackage{soul}
\usepackage{url}
\usepackage{booktabs}
\usepackage{subcaption}
\usepackage{xcolor}
\usepackage{array,multirow}
\usepackage{colortbl}
\usepackage{subcaption}

\newboolean{include-notes}
\setboolean{include-notes}{true}
\newcommand{\sm}[1]{\ifthenelse{\boolean{include-notes}}
 {{\color{orange}SM: #1}}{}}

\newcommand{\nagents}{N}
\newcommand{\ntimesteps}{T}
\newcommand{\transitionprob}{P}
\newcommand{\ntrajectories}{K}
\newcommand{\niters}{M}
\newcommand{\iter}{m}
\newcommand{\dataset}{\mathcal{D}}
\newcommand{\aindex}{i}
\newcommand{\rew}{r}
\newcommand{\allstateinfo}{x}
\newcommand{\allstateinfoset}{\mathcal{X}}

\newcommand{\act}{a}
\newcommand{\jointact}{\Vec{\act}}

\newcommand{\ob}{o}
\newcommand{\obsspace}{\mathcal{O}}
\newcommand{\actspace}{\mathcal{A}}
\newcommand{\statespace}{\mathcal{S}}
\newcommand{\s}{s} 

\newcommand{\exploitability}{\mathcal{E}}
\newcommand{\maximumdepth}{\delta}

\newcommand{\team}{Z}

\newcommand{\STATE}{\State}
\newcommand{\FOR}{\For}
\newcommand{\ENDFOR}{\EndFor}

\newcommand{\agentreward}{\rew_\aindex}
\newcommand{\jointaction}{\overrightarrow{a}}
\newcommand{\agentobsspace}{\obsspace_\aindex}
\newcommand{\allobs}{\obsspace_1, ... , \obsspace_\nagents}
\newcommand{\allacts}{\actspace_1, ..., \actspace_\nagents}
\newcommand{\peragentdataset}{\dataset_\aindex}

\newcommand{\agentob}{\ob_\aindex}
\newcommand{\agentactspace}{\actspace_\aindex}

\newcommand{\maviper}{MAVIPER}
\newcommand{\iviper}{IVIPER}
\newcommand{\ma}{multi-agent}
\newcommand{\dt}{DT}

\newcommand{\imitationbaseline}{Imitation DT}

\newcommand{\maxtreedepth}{6}

\begin{document}
\title{MAVIPER: Learning Decision Tree Policies \\ for Interpretable Multi-Agent \\ Reinforcement Learning}

%
\titlerunning{MAVIPER: Learning Decision-Tree Policies for Interpretable MARL}
%

\author{
Stephanie Milani$^*$\inst{1}{\Letter} \and
Zhicheng Zhang$^*$\inst{2} \and \\
Nicholay Topin\inst{1} \and 
Zheyuan Ryan Shi\inst{1} \and 
Charles Kamhoua\inst{3} \and \\
Evangelos E. Papalexakis\inst{4} \and 
Fei Fang\inst{1}
}
\authorrunning{S. Milani$^*$ and Z. Zhang$^*$ et al.}
%
\tocauthor{Stephanie Milani$^*$,
Zhicheng Zhang$^*$,
Nicholay Topin, 
Zheyuan Ryan Shi,
Charles Kamhoua,
Evangelos E. Papalexakis, 
Fei Fang}
\toctitle{MAVIPER: Learning Decision Tree Policies for Interpretable Multi-Agent Reinforcement Learning}

\institute{
Carnegie Mellon University \email{smilani@andrew.cmu.edu} \and
Shanghai Jiao Tong University \and
Army Research Lab \and 
University of California, Riverside \\
}

\maketitle              
\def\thefootnote{*}\footnotetext{Equal contribution}

\begin{abstract}
Many recent breakthroughs in \ma~reinforcement learning (MARL) require the use of deep neural networks, which are challenging for human experts to interpret and understand.
On the other hand, existing work on interpretable reinforcement learning (RL) has shown promise in extracting more interpretable decision tree-based policies from neural networks, but only in the single-agent setting.
To fill this gap, we propose the first set of algorithms that extract interpretable decision-tree policies from neural networks trained with MARL. 
The first algorithm, \iviper, extends VIPER, a recent method for single-agent interpretable RL, to the \ma{} setting.
We demonstrate that \iviper{} learns high-quality decision-tree policies for each agent. 
To better capture coordination between agents, we propose a novel centralized decision-tree training algorithm, \maviper. 
\maviper{} jointly grows the trees of each agent by predicting the behavior of the other agents using their anticipated trees, and uses resampling to focus on states that are critical for its interactions with other agents. 
We show that both algorithms generally outperform the baselines and that \maviper{}-trained agents achieve better-coordinated performance than \iviper{}-trained agents on three different multi-agent particle-world environments.

\end{abstract}

\section{Introduction}
\label{sec:introduction}
Multi-agent reinforcement learning (MARL) is a promising technique for solving challenging problems, such as air traffic control~\cite{brittain2019autonomous}, train scheduling~\cite{mohanty2020flatland}, cyber defense~\cite{malialis2015distributed}, 
and autonomous driving~\cite{bhalla2020deep}. 
In many of these scenarios, we want to train a \textit{team} of cooperating agents.
Other settings, 
like cyber defense, involve an adversary or set of adversaries with goals that may be at odds with the team of defenders. 
To obtain high-performing agents, most of the recent breakthroughs in MARL rely on neural networks (NNs)~\cite{foerster2017stabilising,rashid2018qmix}, which have thousands to millions of parameters and are challenging for a person to interpret and verify.
Real-world risks necessitate learning \textit{interpretable} policies that people can inspect and verify before deployment, while still performing well at the specified task and being robust to a variety of attackers (if applicable).

Decision trees~\cite{quinlan1986induction} (\dt s) are generally considered to be an \textit{intrinsically} interpretable model family~\cite{molnar2020interpretable}: sufficiently small trees can be contemplated by a person at once (simulatability), have subparts that can be intuitively explained (decomposability), and are verifiable (algorithmic transparency)~\cite{lipton2018mythos}.
In the RL setting, \dt-like models have been successfully used to model transition functions~\cite{strehl2007efficient}, reward functions~\cite{degris2006learning}, value functions~\cite{pyeatt2001decision,tuyls2002reinforcement}, and policies~\cite{mccallum1997reinforcement}.
Although learning \dt~policies for interpretability has been investigated in the single-agent RL setting~\cite{mccallum1997reinforcement,pyeatt2003reinforcement,roth2019conservative}, it has yet to be explored in the \ma~setting.

To address this gap, we propose two algorithms, \iviper{} and \maviper, which combine ideas from model compression and imitation learning to learn \dt~policies in the multi-agent setting.
Both algorithms extend VIPER~\cite{bastani2018verifiable}, which extracts \dt~policies for single-agent RL. 
\iviper{} and \maviper{} work with most existing NN-based MARL algorithms: the policies generated by these algorithms serve as ``expert policies'' and guide the training of a set of \dt~policies.

The main contributions of this work are as follows.
First, we introduce the \iviper{} algorithm as a novel extension of the single-agent VIPER algorithm to \ma~settings.
Indeed, IVIPER trains \dt~policies that achieve high individual performance in the \ma~setting. 
Second, to better capture coordination between agents, we propose a novel centralized DT training algorithm, \maviper. 
\maviper{} jointly grows the trees of each agent by predicting the behavior of the other agents using their anticipated trees. 
To train each agent's policy, \maviper{} uses a novel resampling scheme to find states that are considered critical for its interactions with other agents. 
We show that \maviper{}-trained agents achieve better coordinated performance than \iviper{}-trained agents on three different multi-agent particle-world environments.

\section{Background and Preliminaries}
\label{sec:background}
We focus on the problem of learning interpretable \dt~policies in the multi-agent setting. 
We first describe the formalism of our multi-agent setting, then discuss \dt~policies and review the single-agent version of VIPER.

\subsection{Markov Games and MARL Algorithms}
\label{sec:background_marl}
In MARL, agents act in an environment defined by a Markov game~\cite{littman1994markov,shapley1953stochastic}.
A Markov game for $\nagents$ agents consists of a set of states $\statespace$ describing all possible configurations for all agents, the initial state distribution $\rho : \statespace \rightarrow [0,1]$, and the set of actions $\allacts$ and observations $\allobs$ for each agent $i \in [\nagents]$.
Each agent aims to maximize its own total expected return $R_i = \sum_{t=0}^\infty \gamma^t \agentreward^t $, where $\gamma$ is the discount factor that weights the relative importance of future rewards.
To do so, each agent selects actions using a policy $\pi_{\theta_i} : \agentobsspace \rightarrow \agentactspace$.
After the agents simultaneously execute their actions $\jointaction$ in the environment, the environment produces the next state according to the state transition function $\transitionprob: \statespace \times \actspace_1 \times ... \times \actspace_\nagents \rightarrow \statespace$.
Each agent $\aindex$ receives reward according to a reward function $\agentreward: \statespace \times \actspace_i \rightarrow \mathbb{R}$ 
and a private observation, consisting of a vector of \textit{features}, correlated with the state $\agentob : \statespace \rightarrow \agentobsspace$.

Given a policy profile $\pi=(\pi_1, ..., \pi_\nagents)$, agent $\aindex$'s value function is defined as:
$V^{\pi}_\aindex (s) = \agentreward + \gamma\sum_{s' \in \statespace} \transitionprob(\s, \pi_1(\ob_1), ..., \pi_{\nagents}(\ob_\nagents), \s')V^{\pi}_i(\s')$
and state-action value function is:
$Q^{\pi}_i(s, a_1, ..., a_N) = \agentreward + \gamma \sum_{s' \in S} \transitionprob(s, a_1, ..., a_\nagents,s')V^{\pi}_i(s')$.
We refer to a policy profile excluding agent $\aindex$ as $\pi_{-\aindex}$.

\begin{wrapfigure}{R}{0.49\linewidth}
    \centering
    \includegraphics[width=0.5\textwidth]{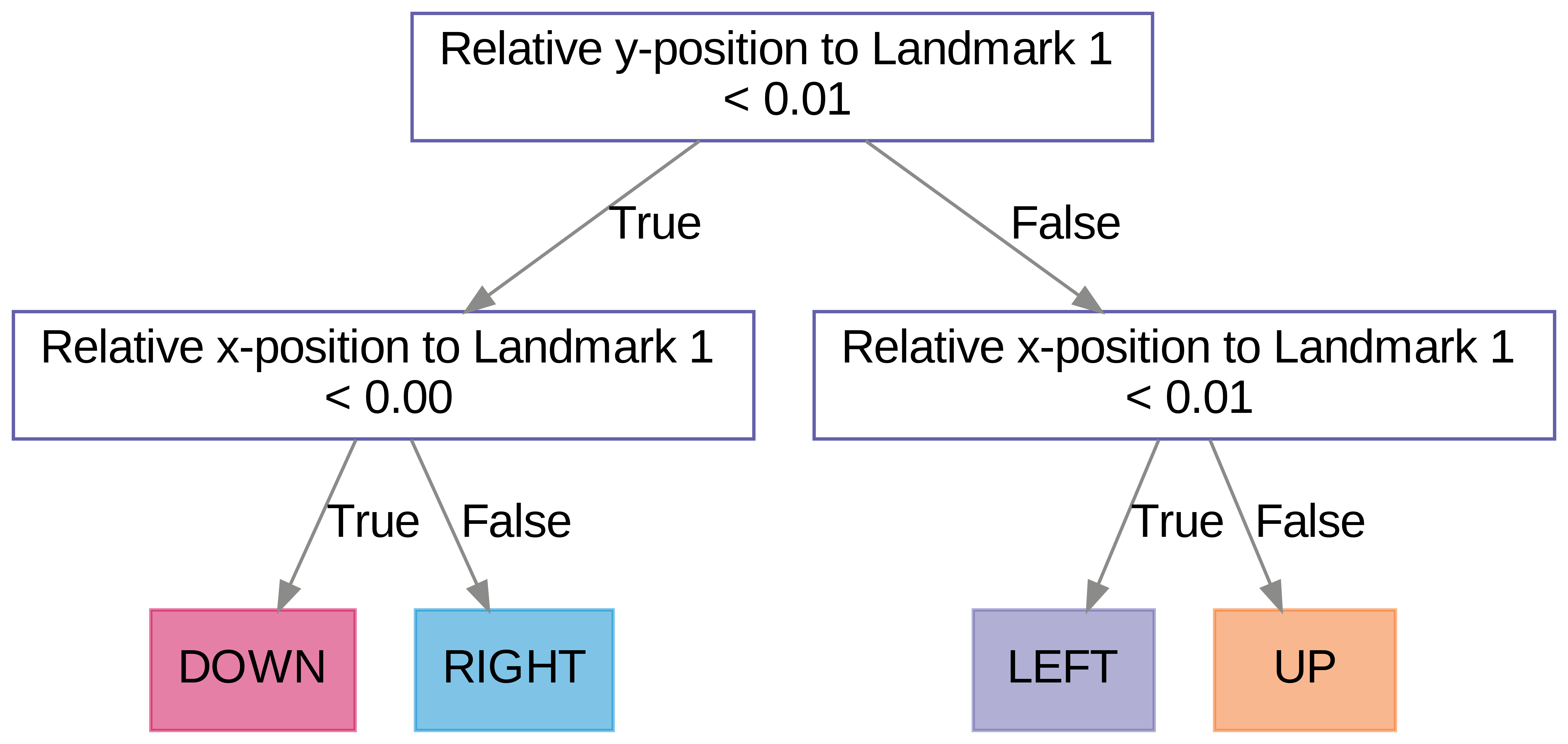}
    \caption{A decision tree of depth two that MAVIPER learns in the Cooperative Navigation environment. The learned decision tree captures the expert's behavior of going to one of the landmarks. }
    \label{fig:example_dt}
\end{wrapfigure}

MARL algorithms fall into two categories: value-based~\cite{rashid2018qmix,son2019qtran,sunehag2017value} and actor-critic~\cite{foerster2018counterfactual,li2019robust,lowe2017multi,yu2021surprising}. 
Value-based methods often approximate $Q$-functions for individual agents in the form of $Q^{\pi}_i(o_i,a_i)$ and derive the policies $\pi_i$ by taking actions with the maximum Q-values. 
In contrast, actor-critic methods often follow the centralized training and decentralized execution (CTDE) paradigm~\cite{oliehoek2008optimal}. 
They train agents in a centralized manner, enabling agents to leverage information beyond their private observation during training; however, agents must behave in a decentralized manner during execution. 
Each agent $\aindex$ uses a centralized critic network $Q^\pi_\aindex$, which takes as input some state information $\allstateinfo$ (including the observations of all agents) and the actions of all agents. 
This assumption addresses the stationarity issue in MARL training: without access to the actions of other agents, the environment appears non-stationary from the perspective of any one agent. 
Each agent $i$ also has a policy network $\pi_i$ that takes as input its observation $\agentob$.


\subsection{Decision Tree Policies}
\label{sec:dtps}

\dt s are tree-like models that recursively partition the input space along a specific feature using a cutoff value.
These models produce axis-parallel partitions:
internal nodes are the intermediate partitions, and leaf nodes are the final partitions.
When used to represent policies, the internal nodes represent the features and values of the input state that the agent uses to choose its action, and the leaf nodes correspond to chosen actions given some input state. 
For an example of a \dt~policy, see Figure~\ref{fig:example_dt}.

\subsection{VIPER}
\label{sec:background_viper}
VIPER~\cite{bastani2018verifiable} is a popular algorithm~\cite{chen2021relace,luss2022local,meng2020interpreting} that extracts \dt~policies for a finite-horizon Markov decision process given an \textit{expert} policy trained using any single-agent RL algorithm. 
It combines ideas from model compression~\cite{bucilua2006model,hinton2015distilling} and imitation learning~\cite{abbeel2004apprenticeship} --- specifically, a variation of the DAGGER algorithm~\cite{ross2011reduction}. 
It uses a high-performing deep NN that approximates the state-action value function to guide the training of a \dt~policy.

VIPER trains a \dt~policy $\hat{\pi}^{m}$ in each iteration $m$; the final output is the best policy among all iterations. 
More concretely, in iteration $m$, it samples $\ntrajectories$ trajectories $\{(s, \hat{\pi}^{\iter-1}(s)) \sim d^{\hat{\pi}^{\iter-1}}\}$ following the \dt~policy trained at the previous iteration. Then, it uses the expert policy $\pi^*$ to suggest actions for each visited state, leading to the dataset $D^m = \{(s, \pi^*(s)) \sim d^{\hat{\pi}^{\iter-1}} \}$ (Line~\ref{algline:vipersample}, Alg.~\ref{alg:viper}). 
VIPER adds these relabeled experiences to a dataset $\dataset$ consisting of experiences from previous iterations. 
Let $V^{\pi^*}$ and $Q^{\pi^*}$ be the state value function and state-action value function given the expert policy $\pi^*$. 
VIPER resamples points $(s, a) \in \dataset$ according to weights: 
    $\tilde{l}(s) = V^{\pi^*} (s) - \min_{a \in A} Q^{\pi^*} (s, a)$.
See~\Cref{alg:viper} in~\Cref{app:alg} for the full VIPER algorithm.


\section{Approach}
\label{sec:approach}
We present two algorithms: \iviper{} and \maviper. Both are general policy extraction algorithms for the \ma{} setting inspired by the single-agent VIPER algorithm.
At a high level, given an expert policy profile $\pi^*=(\pi_1^*, ... \pi_\nagents^*)$ with associated state-action value functions $Q^{\pi^*}=(Q^{\pi^*}_1, ..., Q^{\pi^*}_\nagents)$ trained by an existing MARL algorithm, both algorithms produce a \dt~policy $\hat{\pi}_\aindex$ for each agent $\aindex$. 
These algorithms work with various state-of-art MARL algorithms, including value-based and \ma~actor-critic methods. 
We first discuss \iviper{}, the basic version of our multi-agent \dt{} learning algorithm.
We then introduce additional changes that form the full \maviper{} algorithm.

\subsection{IVIPER}
\label{sec:approach_iviper}
\begin{algorithm}[tb]
\caption{IVIPER in Multi-Agent Setting}
\label{alg:saviper}
\textbf{Input}: 
$(X, A, \transitionprob, R)$, $\pi^*$ 
, $Q^{\pi^*}=(Q^{\pi^*}_1, ..., Q^{\pi^*}_\nagents)$, $\ntrajectories$, $\niters$%
\\
\textbf{Output}: $\hat{\pi}_1, ..., \hat{\pi}_\nagents$ 
\begin{algorithmic}[1] 
\FOR{i=1 to \nagents}
\STATE Initialize dataset $\peragentdataset \gets \emptyset$ and policy $\hat{\pi}_i^0 \gets \pi_i^*$
\FOR{$\iter=1$ to $\niters$}
\STATE Sample $\ntrajectories$ trajectories: $\peragentdataset^\iter \gets \{(x, \pi_1^*(o_1), ..., \pi_\nagents^*(o_\nagents)) \sim d^{\hat{\pi}_i^{\iter-1}, \pi^*_{-i}} \}$ \label{line:sample-trajs}
\STATE Aggregate dataset $\peragentdataset \gets \peragentdataset \cup \peragentdataset^\iter$
\STATE Resample dataset according to loss: 

$\qquad \peragentdataset' \gets \{ (\allstateinfo, \overrightarrow{a}) \sim p((\allstateinfo,\jointaction)) \propto \tilde{l}_i(\allstateinfo) \mathbb{I}[(\allstateinfo,\jointaction) \in \peragentdataset] \}$ \label{line:resample}
\STATE Train decision tree $\hat{\pi}^\iter_i \gets $ TrainDecisionTree($\peragentdataset'$) \label{line:dt-train}
\ENDFOR
\STATE Get best policy $\hat{\pi}_i \gets $ BestPolicy($\hat{\pi}_i^1, ..., \hat{\pi}_i^\niters, \pi_{-i}^*$) \label{line:choose-policy}
\ENDFOR
\STATE \textbf{return} Best policies for each agent $\hat{\pi} = (\hat{\pi}_1, ..., \hat{\pi}_\nagents)$ \label{line:best-policy}
\end{algorithmic}
\end{algorithm}

Motivated by the practical success of single-agent RL algorithms in the MARL setting~\cite{matignon2012independent,berner2019dota}, we extend single-agent VIPER to the multi-agent setting by independently applying the single-agent algorithm to each agent, with a few critical changes described below.
\Cref{alg:saviper} shows the full \iviper{} pseudocode. 

First, we ensure that each agent has sufficient information for training its \dt{} policy.
Each agent has its own dataset $\peragentdataset$ of training tuples.
When using VIPER with multi-agent actor-critic methods that leverage a per-agent centralized critic network $Q^\pi_\aindex$, we ensure that each agent's dataset $\peragentdataset$ has not only its observation and actions, but also the complete state information $\allstateinfo$ --- which consists of the observations of all of the agents --- and the expert-labeled actions of all of the other agents $\pi^*_{j}(o_j) \forall j \neq \aindex$.
By providing each agent with the information about all other agents, we avoid the stationarity issue that arises when the policies of all agents are changing throughout the training process (like in MARL).

Second, we account for important changes that emerge from moving to a multi-agent formalism.
When we sample and relabel trajectories for training each agent's \dt{} policy, 
we sample from the 
distribution $d^{\hat{\pi}_\aindex^{m-1}, \pi^*_{-\aindex}}$ induced by agent $\aindex$'s policy at the previous iteration $\hat{\pi}_\aindex^{m-1}$ and the expert policies of all other agents $\pi^*_{-\aindex}$. 
We only relabel the action for agent $\aindex$ because the other agents choose their actions according to $\pi^*$. 
It is equivalent to treating all other expert agents as part of the environment and only using \dt~policy for agent $\aindex$. 

Third, we incorporate the actions of all agents when resampling the dataset to construct a new, weighted dataset (\Cref{line:resample}, \Cref{alg:saviper}). 
If the MARL algorithm uses a centralized critic $Q(s,\jointaction)$, we resample points 
according to:
\begin{equation}
p((\allstateinfo, a_1, ..., a_\nagents)) \propto \tilde{l}_i(\allstateinfo)
    \mathbb{I}[(\allstateinfo, a_1, ..., a_\nagents) \in \peragentdataset],
    \label{eq:resampling}
\end{equation}
where,
\begin{equation}
    \tilde{l}_i(\allstateinfo) = V^{\pi^*}_i(\allstateinfo) - \min_{a_i \in \agentactspace} Q^{\pi^*}_i(\allstateinfo, a_i, \overrightarrow{a}_{-\aindex}) |_{\jointaction_{-\aindex} = \pi^*_j(o_j) \forall j \neq \aindex}.
    \label{eq:iviper-loss}
\end{equation}
Crucially, we include the actions of all other agents in~\Cref{eq:iviper-loss} to select agent $\aindex$'s minimum Q-value from its centralized state-action value function.

When applied to value-based methods, \iviper~is more similar to single-agent VIPER. 
In particular, in 
\Cref{line:sample-trajs}, \Cref{alg:saviper}, it is sufficient to only store $\agentob$ and $\pi_\aindex^*(\agentob)$ in the dataset $\peragentdataset^\iter$, although we still must sample trajectories according to $\hat{\pi}_\aindex^{m-1}$ and $\pi_{-\aindex}^*$. 
In~\Cref{line:resample}, 
we use $
    \tilde{l}(x) = V^{\pi^*}_\aindex(s) - \min_{a_i \in \agentactspace} Q^{\pi^*}_i (\agentob, a_i)$ from single-agent VIPER, 
removing the reliance of the loss on a centralized critic.

Taken together, these algorithmic changes form the basis of the \iviper{} algorithm.
This algorithm can be viewed as transforming the \ma{} learning problem to a single-agent one, in which other agents are folded into the environment.
This approach works well if
i) we only want an interpretable policy for a single agent in a \ma{} setting or ii) agents do not need to \textit{coordinate} with each other.
When coordination is needed, this algorithm does not reliably capture coordinated behaviors, as each \dt{} is trained independently without consideration for what the other agent's resulting \dt{} policy will learn.
This issue is particularly apparent when trees are constrained to have a small maximum depth, as is desired for interpretability.

\subsection{\maviper} 
\label{sec:approach_maviper}

\begin{algorithm}[ht!]
\caption{\maviper{} (Joint Training)}
\label{alg:maviper}
\textbf{Input}: 
$(\allstateinfoset, A, \transitionprob, R)$, $\pi^*$, $Q^{\pi^*}=(Q^{\pi^*}_1, \dots, Q^{\pi^*}_\nagents)$, $\ntrajectories$, $\niters$ %
\\
\textbf{Output}: $(\hat{\pi}_1, \dots, \hat{\pi}_\nagents)$
\begin{algorithmic}[1] 
\STATE Initialize dataset $\dataset \gets \emptyset $ and policy for each agent $\hat{\pi}_i^0 \gets \pi_i^*\ \forall i \in \nagents$
\FOR{$\iter=1$ to $\niters$}
\STATE Sample $\ntrajectories$ trajectories: $\dataset^\iter \gets \{(\allstateinfo, \pi_1^*(o_1), \dots, \pi_\nagents^*(o_\nagents)) \sim d^{(\hat{\pi}_1^{\iter-1},\dots,\hat{\pi}_\nagents^{\iter-1})} \}$ 
\STATE Aggregate dataset $\dataset \gets \dataset \cup \dataset^\iter$
\STATE For each agent $i$, resample $\mathcal{D}_i$ according to loss: 

$\dataset_i \gets \{ (\allstateinfo, \jointact) \sim p((\allstateinfo,\jointact)) \propto \tilde{l}_i(\allstateinfo) \mathbb{I}[(\allstateinfo,\jointact) \in \dataset] \} \forall i \in \nagents$

\STATE Jointly train \dt s: $(\hat{\pi}_1^\iter,\dots,\hat{\pi}_\nagents^\iter) \gets $ TrainJointTrees($\dataset_1, \dots, \dataset_\nagents$)
\ENDFOR
\State \Return Best set of agents $\hat{\pi} = (\hat{\pi}_1, \dots , \hat{\pi}_\nagents) \in \{ (\hat{\pi}^1_1, \dots, \hat{\pi}^1_\nagents), \dots, (\hat{\pi}^\niters_1, \dots, \hat{\pi}^\niters_\nagents) \}$ 
\item[] 

\Function{TrainJointTrees}{$\dataset_1, \dots, \dataset_\nagents$} \label{alg: maviper_train_joint}
    \State Initialize decision trees $\hat{\pi}^\iter_1,\dots,\hat{\pi}^\iter_\nagents$.
    \Repeat 
        \State Grow one more level for agent $i$'s tree $\hat{\pi}^\iter_i \gets$ Build($\hat{\pi}^\iter_1,\dots,\hat{\pi}^\iter_\nagents$, $\dataset_i$)
        \State Move to the next agent: $i \gets (i + 1) \% \nagents$
    \Until{all trees have grown to the maximum depth allowed}
    \State \Return decision trees $\hat{\pi}^\iter_1,\dots,\hat{\pi}^\iter_\nagents$
\EndFunction

\item[] 

\Function{Build}{$\hat{\pi}^\iter_1,\dots,\hat{\pi}^\iter_\nagents$, $\dataset_i$} \label{alg: maviper_build}
        \For {each data point $ (\allstateinfo, \jointact) \in \dataset_i$}
            \State // Will agent $j$'s (projected) final \dt{} predict its action correctly?
            \State $v_j \gets$ $\mathbb{I}\left[ \text{Predict}(\hat{\pi}^\iter_j, \allstateinfo) = \act_j\right] \forall j\in [1,\nagents] $
            \State // This data point is useful only if many agents' final \dt{}s predict correctly.
            \If {$\sum_{j=1}^\nagents v_j < \text{threshold}$}
                 Remove $d$ from dataset: $\dataset_i \gets \dataset_i \setminus \{(\allstateinfo, \jointact)\}$
            \EndIf
        \EndFor
        \State $\hat{\pi}^\iter_i \gets $ Calculate best next feature split for \dt{} $\hat{\pi}^\iter_i$ using $\dataset_i$. 
    \State \Return $\hat{\pi}^\iter_i$
\EndFunction

\item[]
\Function{Predict}{$\hat{\pi}^\iter_j$, $\allstateinfo$} \label{alg: maviper_predict}
    \State Use $\allstateinfo$ to traverse $\hat{\pi}^\iter_j$ until leaf node $l(x)$
    \State Train a projected final \dt{} $\hat{\pi}'_j \gets $ TrainDecisionTree($\dataset_{j}$)
    \State \Return $\pi$.predict($\allstateinfo$)
\EndFunction
\end{algorithmic}
\end{algorithm}

To address the issue of coordination, we propose \maviper, our novel algorithm for centralized training of coordinated \ma{} \dt{} policies. For expository purpose, we describe \maviper{} in a fully cooperative setting, then explain how to use \maviper{} for mixed cooperative-competitive settings.
At a high-level, \maviper{} trains all of the \dt{} policies, one for each agent, in a centralized manner. 
It jointly grows the trees of each agent by predicting the behavior of the other agents in the environment using their anticipated trees. 
To train each \dt{} policy, \maviper{} employs a new resampling technique to find states that are critical for its interactions with other agents. 
\Cref{alg:maviper} shows the full \maviper{} algorithm.
Specifically, \maviper{} is built upon the following extensions to \iviper{} that aim at addressing the issue of coordination.

First, \maviper{} does not calculate the probability $p(\allstateinfo)$ of a joint observation $\allstateinfo$ by viewing the other agents as stationary experts. 
Instead, MAVIPER focuses on the critical states where a good joint action can make a difference. 
Specifically, MAVIPER aims to measure how much worse off agent $i$ would be, taking expectation over all possible joint actions of the other agents, if it acts in the worst way possible compared with when it acts in the same way as the expert agent.
So, we define $l_i(\allstateinfo)$, as in~\Cref{eq:iviper-loss}, as:
\begin{equation}
    \tilde{l}_i(\allstateinfo) = \mathbb{E}_{\jointaction_{-\aindex}}\left[Q^{\pi^*}_i\left(\allstateinfo, \pi^*_i(x), \jointaction_{-\aindex}\right) - 
    \min_{a_i \in \agentactspace} Q^{\pi^*}_i\left(\allstateinfo, a_i, \overrightarrow{a}_{-\aindex}\right)\right].
    \label{eq:maviper-loss}
\end{equation}
\maviper{} uses the DT policies $(\hat{\pi}_1^{m-1},\dots,\hat{\pi}_N^{m-1})$ from the last iteration to perform rollouts and collect new data. 

Second, we add a prediction module to the DT training process to increase the \textit{joint} accuracy, as shown in the $\texttt{Predict}$ function. 
The goal of the prediction module is to predict the actions that the other \dt s $\{\hat{\pi}_j\}_{j \neq i}$ might make, given their partial observations.
To make the most of the prediction module, \maviper{} grows the trees evenly using a breadth-first ordering to avoid biasing towards the result of any specific tree.
Since the trees are not complete at the time of prediction, we use the output of another \dt{} trained with the full dataset associated with that node for the prediction.
Following the intuition that the correct prediction of one agent alone may not yield much benefit if the other agents are wrong, we use this prediction module to remove all data points whose proportion of correct predictions is lower than a predefined threshold.
We then calculate the splitting criteria based on this modified dataset and continue iteratively growing the tree.

In some mixed cooperative-competitive settings, agents in a team share goals and need to coordinate with each other, but they face other agents or other teams whose goals are not fully aligned with theirs. In these settings, MAVIPER follows a similar procedure to jointly train policies for agents in the same team to ensure coordination. More specifically, for a team $\team$, the $\texttt{Build}$ and $\texttt{Predict}$ function is constrained to only make predictions for the agents in the same team. \Cref{eq:maviper-loss} now takes the expectation over the joint actions for agents outside the team and becomes:
\begin{equation}
\tilde{l}_i(\allstateinfo) = \mathbb{E}_{\jointaction_{-\team}}\left[Q^{\pi^*}_i\left(\allstateinfo, \pi^*_i(x), \jointaction_{-\team}\right) - 
    \min_{a_i \in \agentactspace} Q^{\pi^*}_i\left(\allstateinfo, a_i, \overrightarrow{a}_{-\team}\right)\right].
\end{equation}

Taken together, these changes comprise the \maviper{} algorithm.
Because we explicitly account for the anticipated behavior of other agents in both the predictions and the resampling probability, we hypothesize that \maviper{} will better capture coordinated behavior.

\section{Experiments}
\label{sec:experiments}

We now investigate how well \maviper{} and \iviper{} agents perform in a variety of environments.
Because the goal is to learn high-performing yet interpretable policies, we evaluate the quality of the trained policies in three multi-agent environments: two mixed competitive-cooperative environments and one fully cooperative environment.
We measure how well the DT policies perform in the environment because our goal is to \textit{deploy} these policies, not the expert ones.

Since small DTs are considered interpretable, we constrain the maximum tree depth to be at most $\maxtreedepth$.
The expert policies used to guide the \dt{} training are generated by MADDPG~\cite{lowe2017multi}
\footnote{We use the Pytorch~\cite{paszke2017automatic} implementation \url{https://github.com/shariqiqbal2810/maddpg-pytorch}.}.
We compare to two baselines:
\begin{enumerate}
    \item Fitted Q-Iteration. We iteratively approximate the Q-function with a regression \dt{}~\cite{ernst2005tree}. We discretize states to account for continuous state values. More details in~\Cref{app:imp}. We derive the policy by taking the the action associated with the highest estimated Q-value for that input state.
    \item \imitationbaseline. Each \dt{} policy is directly trained using a dataset collected by running the expert policies for multiple episodes. No resampling is performed. The observations for an agent are the features, and the actions for that agent are the labels.
\end{enumerate}
We detail the hyperparameters and the hyperparameter-selection process in \Cref{app:hyperparams}. 
We train a high-performing MADDPG expert, then run each \dt-learning algorithm $10$ times with different random seeds.
We evaluate all policies 
by running $100$ episodes.
Error bars correspond to the $95 \%$ confidence interval.
Our code is available through our project website: \url{https://stephmilani.github.io/maviper/}.

\subsection{Environments}
\label{sec:experiments_environments}
We evaluate our algorithms on three multi-agent particle world environments~\cite{lowe2017multi}, described below. 
Episodes terminate when the maximum number of timesteps $\ntimesteps = 25$ is reached.
We choose the primary performance metric based on the environment (detailed below), and we also provide results using expected return as the performance metric in~\Cref{app:results}.


\paragraph{Physical Deception.}


In this environment, 
a team of $\nagents$ defenders must protect $\nagents$ targets from one adversary.
One of the targets is the true target, which is known to the defenders but not to the adversary.
For our experiments, $\nagents = 2$.
Defenders succeed during an episode if they split up to cover all of the targets simultaneously; the adversary succeeds if it reaches the true target during the episode. 
Covering and reaching targets is defined as being $\epsilon$-close to a target for at least one timestep during the episode.  
We use the defenders' and the adversary's success rate as the primary performance metric in this environment.

\paragraph{Cooperative Navigation.}
This environment consists of a team of $\nagents$ agents, who must learn to cover all $\nagents$ targets while avoiding collisions with each other.
For our experiments, $\nagents=3$.
Agents succeed during an episode if they split up to cover all of the targets without colliding.
Our primary performance metric is the summation of the distance of the closest agent to each target, for all targets. 
Low values of the metric indicate that the agents correctly learn to split up.

\paragraph{Predator-prey.}
This variant involves a team of $K$ slower, cooperating predators that chase $M$ faster prey.
There are $L=2$ landmarks impeding the way.
We choose $K=M=2$. 
We assume that each agent has a restricted observation space mostly consisting of binarized relative positions and velocity (if applicable) of the landmarks and other agents in the environment. 
See \Cref{app:env} for full details.
Our primary performance metric is the number of collisions between predators and prey.
For prey, lower is better; for predators, higher is better.

\subsection{Results}
\label{sec:results}

For each environment, we compare the \dt{} policies generated by different methods and check if \iviper{} and \maviper{} agents achieve better performance ratio than the baselines overall.
We also investigate whether \maviper{} learns better coordinated behavior than \iviper{}.
Furthermore, we investigate which algorithms are the most robust to different types of opponents.
We conclude with an ablation study to determine which components of the \maviper{} algorithm contribute most to its success.

\subsubsection{Individual Performance Compared to Experts} 

\begin{figure}[t!]
    \centering 
    \begin{subfigure}[b]{.65\textwidth}
        \centering 
        \begin{subfigure}[b]{0.49\textwidth}
        \includegraphics[width=1.0\textwidth]{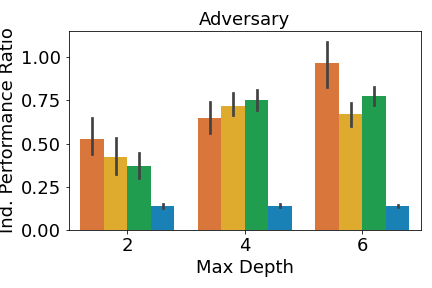}
        \end{subfigure}
        \centering
        \begin{subfigure}[b]{0.49\textwidth}
        \includegraphics[width=1.0\textwidth]{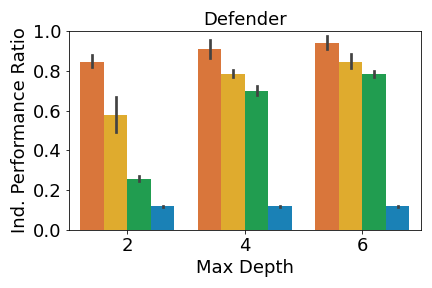}
        \end{subfigure}
        \caption{Physical Deception}
        \label{fig:pd_indiv}
    \end{subfigure}
    \begin{subfigure}[b]{0.32\textwidth}
        \includegraphics[width=1.0\textwidth]{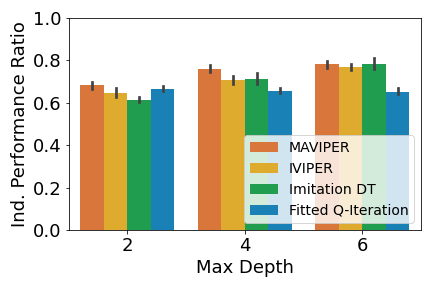}
        \caption{Cooperative Navigation}
        \label{fig:cn_indiv}
    \end{subfigure}
    \begin{subfigure}[b]{1.0\textwidth}
        \centering 
        \begin{subfigure}[b]{0.33\textwidth}
        \includegraphics[width=1.0\textwidth]{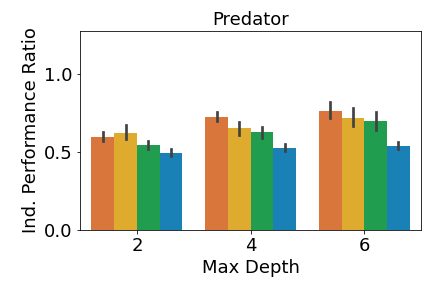}
        \end{subfigure}
        \centering
        \begin{subfigure}[b]{0.33\textwidth}
        \includegraphics[width=1.0\textwidth]{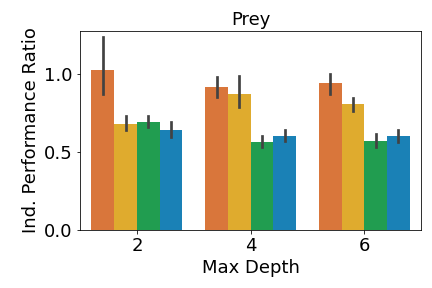}
       \end{subfigure}
        \caption{Predator-prey}
        \label{fig:pp_indiv}
    \end{subfigure}
    \caption{
    Individual performance ratio: Relative performance when only one agent adopts \dt~policy and all other agents use expert policy. 
    } 
    \label{fig:indiv}
\end{figure}

We analyze the performance of the \dt~policies when only one agent adopts the \dt~policy while all other agents use the expert policies. Given a \dt~policy profile $\hat{\pi}$ and the expert policy profile $\pi^*$, if agent $i$ who belongs to team $\team$ uses its \dt~policy, then the individual performance ratio is defined as:
$\frac{U_\team(\hat{\pi}_i, \pi^*_{-i})}{U_\team(\pi^*)}$,
where $U_\team(\cdot)$ is team $\team$'s performance given the agents' policy profile (since we define our primary performance metrics at the team level).
A performance ratio of $1$ means that the \dt~policies perform as well as the expert ones. We can get a ratio above $1$, since we compare the performance of the \dt{} and the expert policies in the environment, not the similarity of the \dt{} and expert policies.

We report the mean individual performance ratio for each team in Figure~\ref{fig:indiv}, averaged over all trials and all agents in the team. 
As shown in~\Cref{fig:pd_indiv}, individual \maviper{} and \iviper{} defenders outperform the two baselines for all maximum depths in the physical deception environment. 
However, \maviper{} and \iviper{} adversaries perform similarly to the \imitationbaseline{} adversary, indicating that the correct strategy may be simple enough to capture with a less-sophisticated algorithm.
Agents also perform similarly on the cooperative navigation environment (\Cref{fig:cn_indiv}). 
As mentioned in the original MADDPG paper~\cite{lowe2017multi}, this environment has a less stark contrast between success and failure, so these results are not unexpected.

In predator-prey, we see the most notable performance difference when comparing the predator. 
When the maximum depth is 2, only \maviper{} achieves near-expert performance.
When the maximum depths are 4 and 6, \maviper{} and \iviper{} agents achieve similar performance and significantly outperform the baselines.
The preys achieve similar performance across all algorithms.
We suspect that the complexity of this environment makes it challenging to replace even a single prey's policy with a \dt.

Furthermore, MAVIPER achieves a performance ratio above 0.75 in all environments with a maximum depth of 6. 
The same is true for IVIPER, except for the adversaries in physical deception. 
That means \dt{} policies generated by IVIPER and MAVIPER lead to a performance degradation of less than or around 20\% compared to the less interpretable NN-based expert policies. These results show that IVIPER and MAVIPER generate reasonable \dt~policies and outperform the baselines overall when adopted by a single agent.

\subsubsection{Joint Performance Compared to Experts}
\begin{figure}[t!]
    \centering 
    \begin{subfigure}[b]{0.4\textwidth}
        \includegraphics[width=1.0\columnwidth]{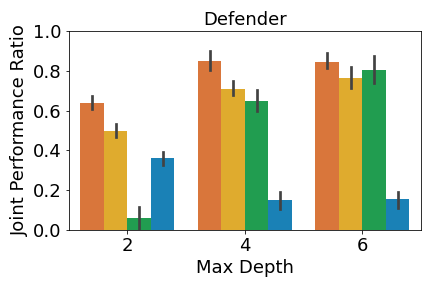}
        \caption{Physical Deception}
        \label{fig:pd_joint}
    \end{subfigure}
    \begin{subfigure}[b]{0.4\textwidth}
        \includegraphics[width=1.0\textwidth]{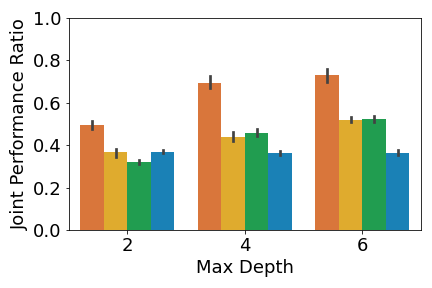}
        \caption{Cooperative Navigation}
        \label{fig:cn_joint}
    \end{subfigure}
    \begin{subfigure}[b]{1.0\textwidth}
        \centering 
        \begin{subfigure}[b]{0.4\textwidth}
        \includegraphics[width=1.0\textwidth]{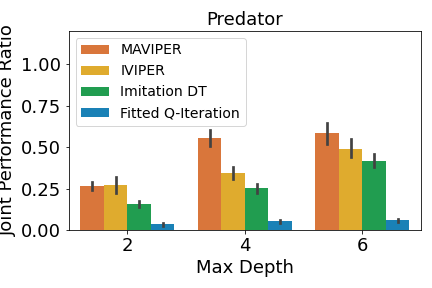}
        \end{subfigure}
        \centering
        \begin{subfigure}[b]{0.4\textwidth}
        \includegraphics[width=1.0\textwidth]{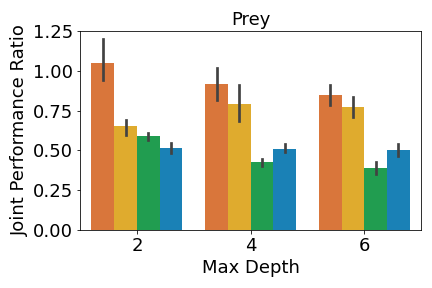}
       \end{subfigure}
        \caption{Predator-prey}
        \label{fig:pp_joint}
    \end{subfigure}
    \caption{
    Joint performance ratio: Relative performance when all agents in a team adopt \dt~policy and other agents use expert policy. 
    } 
    \label{fig:joint}
\end{figure}
A crucial aspect in multi-agent environments is agent coordination, especially when agents are on the same team with shared goals. 
To ensure that the \dt{} policies capture this coordination, we analyze the performance of the \dt{} policies when all agents in a team adopt \dt{} policies, while other agents use expert policies.
We define the joint performance ratio as:
$\frac{U_\team(\hat{\pi}_\team, \pi^*_{-Z})}{U_\team(\pi^*)}$,
where $U_\team(\hat{\pi}_\team, \pi^*_{-Z})$ is the utility of team $\team$ when using their \dt{} policies against the expert policies of the other agents $-\team$.
\Cref{fig:joint} shows the mean joint performance ratio for each team, averaged over all trials. 

\begin{wrapfigure}{R}{0.49\linewidth}
    \centering 
    \includegraphics[width=.49\textwidth]{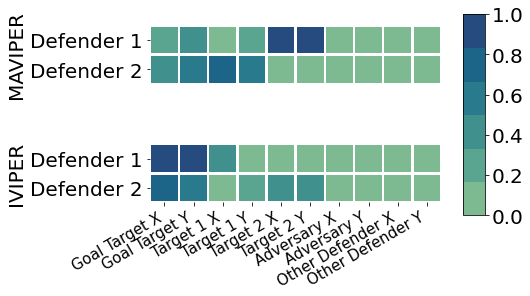}
    \caption{Features used by the two defenders in the physical deception environment. Actual features are the relative positions of that agent and the labeled feature. Darker squares correspond to higher feature importance. \maviper{} defenders most commonly split importance across the two targets.} 
    \label{fig:feature_importances}
\end{wrapfigure}
\Cref{fig:pd_joint} shows that \maviper{} defenders outperform \iviper{} and the baselines, indicating that it better captures the coordinated behavior necessary to succeed in this environment. 
Fitted Q-Iteration struggles to achieve coordinated behavior, despite obtaining non-zero success for individual agents.
This algorithm cannot capture the coordinated behavior, which we suspect is due to poor Q-value estimates.
We hypothesize that the superior performance of \maviper{} is partially due to the defender agents correctly splitting their ``attention'' to the two targets to induce the correct behavior of covering both targets.
To investigate this, we inspect the normalized average feature importances of the \dt{} policies of depth $4$ for both \iviper{} and \maviper{} over $5$ of the trials, as shown in \Cref{fig:feature_importances}.
Each of the \maviper{} defenders (top) most commonly focuses on the attributes associated with one of the targets. More specifically, defender 1 focuses on target 2 and defender 2 focuses on target 1. 
In contrast, both \iviper{} defenders (bottom) mostly focus on the attributes associated with the goal target.
Not only does this overlap in feature space mean that defenders are unlikely to capture the correct covering behavior, but it also leaves them more vulnerable to an adversary, as it is easier to infer the correct target.

\Cref{fig:cn_joint} shows that \maviper{} agents significantly outperform all other algorithms in the cooperative navigation environment for all maximum depths.
\iviper{} agents significantly outperform the baselines for a maximum depth of $2$ but achieve similar performance to the \imitationbaseline{} for the other maximum depths (where both algorithms significantly outperform the Fitted Q-Iteration baseline).
\maviper{} better captures coordinated behavior, even as we increase the complexity of the problem by introducing another cooperating agent.

\Cref{fig:pp_joint} shows that the prey teams trained by \iviper{} and MAVIPER outperform the baselines for all maximum depths.
The predator teams trained by \iviper{} and \maviper{} similarly outperform the baselines for all maximum depths. Also, \maviper{} leads to better performance than \iviper{} in two of the settings (prey with depth 2 and predator with depth 4) while having no statistically significant advantage in other settings.
Taken together, these results indicate that \iviper{} and \maviper{} better capture the coordinated behavior necessary for a team to succeed in different environments, with \maviper{} significantly outperforming \iviper{} in several environments.


\subsubsection{Robustness to Different Opponents}
\begin{table}[t]
    \centering
    \begin{tabular}{cccccc}
        Environment & Team & \maviper & \iviper & Imitation & Fitted  \\
         &  & & & DT & Q-Iteration \\
        \toprule 
        Physical & Defender & \textbf{.77} (.01) & .33 (.01) & .24 (.03) & .004 (.00) \\
        Deception & Adversary & \textbf{.42} (.03) & \textbf{.41} (.03) & \textbf{.42} (.03) & .07 (.01) \\        
        \midrule 
        Predator-& Predator & \textbf{2.51} (0.72) & \textbf{1.98} (0.58) & 1.14 (0.28) & 0.26 (0.11) \\
        prey & Prey & 1.76 (0.80) & \textbf{2.16} (1.24) & \textbf{2.36} (1.90) & 1.11 (0.82) \\
        \bottomrule
    \end{tabular}
    \caption{Robustness results. We report mean team performance and standard deviation of \dt~policies for each team, averaged across a variety of opponent policies. The best-performing algorithm for each agent type is shown in \textbf{bold}.  
    }
    \label{tab:avg_robustness}
\end{table}
We investigate the robustness of the \dt~policies when a team using \dt~policies plays against a variety of opponents in the mixed competitive-cooperative environments.
For this set of experiments, we choose a maximum depth of $4$. 
Given a \dt~policy profile $\hat{\pi}$, a team \team's performance against an alternative policy file $\pi'$ used by the opponents is:
$U_\team(\hat{\pi}_\team, \pi'_{-Z})$.
We consider a broad set of opponent policies $\pi'$, including the policies generated by \maviper, \iviper, \imitationbaseline, Fitted Q-Iteration, and MADDPG. 
We report the mean team performance averaged over all opponent policies in \Cref{tab:avg_robustness}.
See~\Cref{tab:robustness_pp,tab:robustness_pd} in \Cref{app:results} for the full results.

For physical deception, \maviper{} defenders outperform all other algorithms, with a gap of $0.44$ between its performance and the next-best algorithm, \iviper. 
This result indicates that \maviper{} learns coordinated defender policies that perform well against various adversaries. 
\maviper{}, \iviper{}, and \imitationbaseline{} adversaries perform similarly on average, with a similar standard deviation, which supports the idea that the adversary's desired behavior is simple enough to capture with a less-sophisticated algorithm.
For predator-prey, \maviper{} predators and prey outperform all other algorithms. 
The standard deviation of the performance of all algorithms is high due to this environment's complexity.

\subsubsection{Ablation Study} 
\begin{figure}[t!]
    \centering 
    \begin{subfigure}[b]{0.32\textwidth}
        \includegraphics[width=1.0\textwidth]{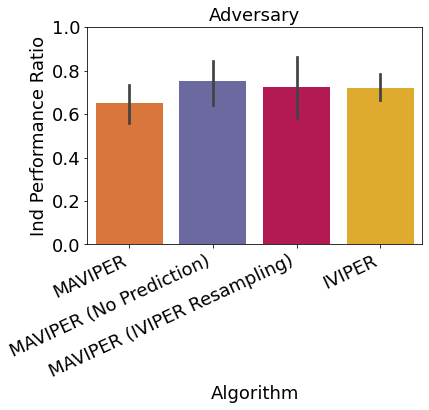}
    \end{subfigure}
        \centering
        \begin{subfigure}[b]{0.32\textwidth}
        \includegraphics[width=1.0\textwidth]{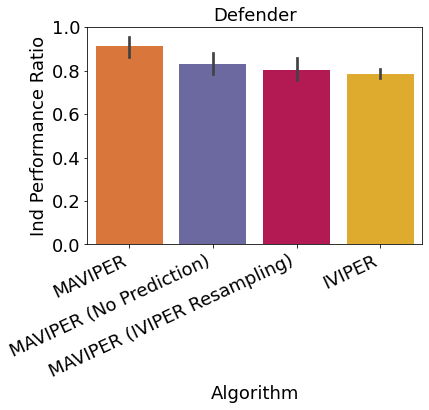}
        \end{subfigure}
        \begin{subfigure}[b]{0.32\textwidth}
    \includegraphics[width=1.0\textwidth]{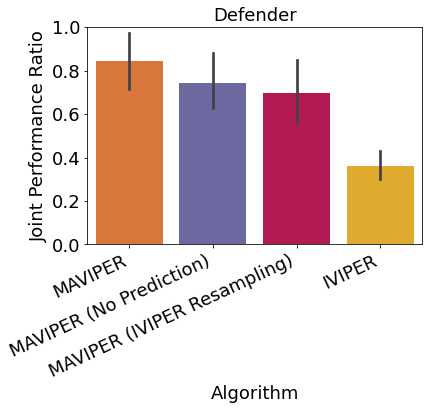}
    \end{subfigure}
    \caption{Ablation study for \maviper{} for a maximum depth of $4$. MAVIPER (No Prediction) does not utilize the predicted behavior of the anticipated \dt s of the other agents to grow each agent's tree. MAVIPER (IVIPER Resampling) uses the same resampling method as IVIPER.}
    \label{fig:ablation}
\end{figure}
As discussed in \Cref{sec:approach_maviper}, \maviper{} improves upon \iviper{} with a few critical changes.
First, we utilize the predicted behavior of the anticipated \dt s of the other agents to grow each agent's tree.
Second, we alter the resampling probability to incorporate the average Q-values over all actions for the other agents. 
To investigate the contribution of these changes to the performance, we run an ablation study with a maximum depth of $4$ on the physical deception environment. 
We report both the mean independent and joint performance ratios for the defender team in \Cref{fig:ablation}, comparing MAVIPER and IVIPER to two variants of MAVIPER without one of the two critical changes. 
Results show that both changes contributed to the improvement of MAVIPER over IVIPER, especially in the joint performance ratio.

\section{Related Work}
\label{sec:related}
Most work on interpretable RL is in the single-agent setting~\cite{milani2022survey}.
We first discuss techniques that directly learn \dt{} policies. 
CUSTARD~\cite{topin2021iterative} extends the action space of the MDP to contain actions for constructing a \dt, i.e., choosing a feature to branch a node. 
Training an agent in the augmented MDP yields a \dt{} policy for the original MDP while still enabling training using any function approximator, like NNs, during training. 
By redefining the MDP, the learning problem becomes more complex, which is problematic in multi-agent settings where the presence of other agents already complicates the learning problem.
A few other works directly learn \dt~policies~\cite{ernst2005tree,mccallum1997reinforcement,uther2000lumberjack} for single-agent RL but not for the purpose of interpretability. 
Further, these works have custom learning algorithms and cannot utilize a high-performing NN policy to guide training.

VIPER~\cite{bastani2018verifiable} is considered to be a post-hoc \dt-learning method~\cite{bastani2018verifiable}; however, we use it to produce \textit{intrinsically} interpretable policies for deployment. 
MOET~\cite{vasic2019moet} extends VIPER by learning a mixture of \dt~policies trained on different regions of the state space.
The resulting policy is a linear combination of multiple trees with non-axis-parallel partitions of the state.
We find that the performance difference between VIPER and MOET is not significant enough to increase the complexity of the policy structure, which would sacrifice interpretability.

Despite increased interest in interpretable single-agent RL, interpretable MARL is less commonly explored.
One line of work generates explanations from non-interpretable policies.
Some work uses attention~\cite{iqbal2019actor,li2019sparsemaac,motokawa2021mat} to select and focus on critical factors that impact agents in the training process. 
Other work generates explanations as verbal explanations with predefined rules~\cite{wang2020explanation} or Shapley values~\cite{heuillet2022collective}.
The most similar line of work to ours~\cite{kazhdan2020marleme} approximates non-interpretable MARL policies to interpretable ones using the framework of abstract argumentation. 
This work constructs argument preference graphs given manually-provided arguments. 
In contrast, our work does not need these manually-provided arguments for interpretability. 
Instead, we generate \dt~policies.

\section{Discussion and Conclusion}

We proposed \iviper{} and \maviper, the first algorithms, to our knowledge, that train interpretable \dt{} policies for MARL.
We evaluated these algorithms on both cooperative and mixed competitive-cooperative environments.
We showed that they can achieve individual performance of at least 75\% of expert performance in most environment settings and over 90\% in some of them, given a maximum tree depth of 6. 
We also empirically validated that \maviper{} effectively captures coordinated behavior by showing that teams of \maviper-trained agents outperform the agents trained by \iviper{} and several baselines.
We further showed that \maviper{} generally produces more robust agents than the other \dt-learning algorithms.

Future work includes learning these high-quality \dt{} policies from fewer samples, e.g., by using dataset distillation~\cite{wang2018dataset}.
We also note that our algorithms can work in some environments where the experts and \dt s are trained on different sets of features. 
Since \dt s can be easier to learn with a simpler set of features, future work includes augmenting our algorithm with an automatic feature selection component that constructs simplified yet still interpretable features for training the \dt{} policies.

\subsubsection{Acknowledgements}
This material is based upon work supported by the Department of Defense (DoD) through the National Defense Science \& Engineering Graduate (NDSEG) Fellowship Program. 
This research was sponsored by the U.S. Army Combat Capabilities Development Command Army Research Laboratory and was accomplished under Cooperative Agreement Number W911NF-13-2-0045 (ARL Cyber Security CRA). Any opinions, findings and conclusions or recommendations expressed in this material are those of the author(s) and do not reflect the views of the funding agencies or government agencies. The U.S. Government is authorized to reproduce and distribute reprints for Government purposes notwithstanding any copyright notation here on.

%
%
 \bibliographystyle{splncs04}
 \bibliography{relworkshortest}

\begin{thebibliography}{10}
\providecommand{\url}[1]{\texttt{#1}}
\providecommand{\urlprefix}{URL }
\providecommand{\doi}[1]{https://doi.org/#1}

\bibitem{abbeel2004apprenticeship}
Abbeel, P., Ng, A.: Apprenticeship learning via inverse reinforcement learning.
  In: ICML (2004)

\bibitem{bastani2018verifiable}
Bastani, O., et~al.: Verifiable reinforcement learning via policy extraction.
  In: NeurIPS (2018)

\bibitem{berner2019dota}
Berner, C., et~al.: Dota 2 with large scale deep reinforcement learning. arXiv
  preprint 1912.06680  (2019)

\bibitem{bhalla2020deep}
Bhalla, S., et~al.: Deep multi agent reinforcement learning for autonomous
  driving. In: Canadian Conf. Artif. Intell. (2020)

\bibitem{brittain2019autonomous}
Brittain, M., Wei, P.: Autonomous air traffic controller: A deep multi-agent
  reinforcement learning approach. arXiv preprint arXiv:1905.01303  (2019)

\bibitem{bucilua2006model}
Buciluǎ, C., et~al.: Model compression. In: KDD (2006)

\bibitem{chen2021relace}
Chen, Z., et~al.: Relace: Reinforcement learning agent for counterfactual
  explanations of arbitrary predictive models. arXiv preprint arXiv:2110.11960
  (2021)

\bibitem{degris2006learning}
Degris, T., et~al.: Learning the structure of factored {M}arkov decision
  processes in reinforcement learning problems. In: ICML (2006)

\bibitem{ernst2005tree}
Ernst, D., et~al.: Tree-based batch mode reinforcement learning. JMLR
  \textbf{6} (2005)

\bibitem{foerster2017stabilising}
Foerster, J., et~al.: Stabilising experience replay for deep multi-agent
  reinforcement learning. In: ICML (2017)

\bibitem{foerster2018counterfactual}
Foerster, J., et~al.: Counterfactual multi-agent policy gradients. In: AAAI
  (2018)

\bibitem{heuillet2022collective}
Heuillet, A., et~al.: Collective explainable ai: Explaining cooperative
  strategies and agent contribution in multiagent reinforcement learning with
  shapley values. IEEE Comput. Intell. Magazine  \textbf{17} (2022)

\bibitem{hinton2015distilling}
Hinton, G., et~al.: Distilling the knowledge in a neural network. arXiv
  preprint arXiv:1503.02531  (2015)

\bibitem{iqbal2019actor}
Iqbal, S., Sha, F.: Actor-attention-critic for multi-agent reinforcement
  learning. In: ICML (2019)

\bibitem{kazhdan2020marleme}
Kazhdan, D., et~al.: Marleme: A multi-agent reinforcement learning model
  extraction library. In: IJCNN (2020)

\bibitem{li2019robust}
Li, S., et~al.: Robust multi-agent reinforcement learning via minimax deep
  deterministic policy gradient. In: AAAI (2019)

\bibitem{li2019sparsemaac}
Li, W., et~al.: Sparsemaac: Sparse attention for multi-agent reinforcement
  learning. In: Int. Conf. Database Syst. for Adv. Appl. (2019)

\bibitem{lipton2018mythos}
Lipton, Z.: The mythos of model interpretability. ACM Queue  \textbf{16}(3)
  (2018)

\bibitem{littman1994markov}
Littman, M.: Markov games as a framework for multi-agent reinforcement
  learning. In: Mach. Learning (1994)

\bibitem{lowe2017multi}
Lowe, R., et~al.: Multi-agent actor-critic for mixed cooperative-competitive
  environments. arXiv preprint arXiv:1706.02275  (2017)

\bibitem{luss2022local}
Luss, R., et~al.: Local explanations for reinforcement learning. arXiv preprint
  arXiv:2202.03597  (2022)

\bibitem{malialis2015distributed}
Malialis, K., Kudenko, D.: Distributed response to network intrusions using
  multiagent reinforcement learning. Eng. Appl. Artif. Intell.  (2015)

\bibitem{matignon2012independent}
Matignon, L., et~al.: Independent reinforcement learners in cooperative markov
  games: a survey regarding coordination problems. Knowledge Eng. Review
  \textbf{27}(1) (2012)

\bibitem{mccallum1997reinforcement}
McCallum, R.: Reinforcement learning with selective perception and hidden
  state. PhD Thesis, Univ. Rochester, Dept. of Comp. Sci.  (1997)

\bibitem{meng2020interpreting}
Meng, Z., et~al.: Interpreting deep learning-based networking systems. In:
  Proceedings of the Annual conference of the ACM Special Interest Group on
  Data Communication on the applications, technologies, architectures, and
  protocols for computer communication (2020)

\bibitem{milani2022survey}
Milani, S., et~al.: A survey of explainable reinforcement learning. arXiv
  preprint arXiv:2202.08434  (2022)

\bibitem{mohanty2020flatland}
Mohanty, S., et~al.: Flatland-rl: Multi-agent reinforcement learning on trains.
  arXiv preprint arXiv:2012.05893  (2020)

\bibitem{molnar2020interpretable}
Molnar, C.: Interpretable Machine Learning (2019)

\bibitem{motokawa2021mat}
Motokawa, Y., Sugawara, T.: Mat-dqn: Toward interpretable multi-agent deep
  reinforcement learning for coordinated activities. In: ICANN (2021)

\bibitem{oliehoek2008optimal}
Oliehoek, F., et~al.: Optimal and approximate q-value functions for
  decentralized pomdps. JAIR  \textbf{32} (2008)

\bibitem{paszke2017automatic}
Paszke, A., et~al.: Automatic differentiation in pytorch  (2017)

\bibitem{pyeatt2003reinforcement}
Pyeatt, L.: Reinforcement learning with decision trees. In: Appl. Informatics
  (2003)

\bibitem{pyeatt2001decision}
Pyeatt, L., Howe, A.: Decision tree function approximation in reinforcement
  learning. In: Int. Symp. on Adaptive Syst.: Evol. Comput. and Prob. Graphical
  Models (2001)

\bibitem{quinlan1986induction}
Quinlan, J.: Induction of decision trees. Mach. Learning  (1986)

\bibitem{rashid2018qmix}
Rashid, T., et~al.: Qmix: Monotonic value function factorisation for deep
  multi-agent reinforcement learning. In: ICML (2018)

\bibitem{ross2011reduction}
Ross, S., et~al.: A reduction of imitation learning and structured prediction
  to no-regret online learning. In: AISTATS (2011)

\bibitem{roth2019conservative}
Roth, A., et~al.: Conservative q-improvement: Reinforcement learning for an
  interpretable decision-tree policy. arXiv preprint arXiv:1907.01180  (2019)

\bibitem{shapley1953stochastic}
Shapley, L.: Stochastic games. PNAS  \textbf{39}(10) (1953)

\bibitem{son2019qtran}
Son, K., et~al.: Qtran: Learning to factorize with transformation for
  cooperative multi-agent reinforcement learning. arXiv preprint
  arXiv:1905.05408  (2019)

\bibitem{strehl2007efficient}
Strehl, A., et~al.: Efficient structure learning in factored-state mdps. In:
  AAAI (2007)

\bibitem{sunehag2017value}
Sunehag, P., et~al.: Value-decomposition networks for cooperative multi-agent
  learning. arXiv preprint arXiv:1706.05296  (2017)

\bibitem{topin2021iterative}
Topin, N., et~al.: Iterative bounding mdps: Learning interpretable policies via
  non-interpretable methods. In: AAAI (2021)

\bibitem{tuyls2002reinforcement}
Tuyls, K., et~al.: Reinforcement learning in large state spaces. In: Robot
  Soccer World Cup (2002)

\bibitem{uther2000lumberjack}
Uther, W., Veloso, M.: The lumberjack algorithm for learning linked decision
  forests. In: Int. Symp. Abstract., Reformulation, and Approx. (2000)

\bibitem{vasic2019moet}
Vasic, M., et~al.: Mo{\"e}t: Interpretable and verifiable reinforcement
  learning via mixture of expert trees. arXiv preprint arXiv:1906.06717  (2019)

\bibitem{wang2018dataset}
Wang, T., et~al.: Dataset distillation. arXiv preprint arXiv:1811.10959  (2018)

\bibitem{wang2020explanation}
Wang, X., et~al.: Explanation of reinforcement learning model in dynamic
  multi-agent system. arXiv preprint arXiv:2008.01508  (2020)

\bibitem{yu2021surprising}
Yu, C., et~al.: The surprising effectiveness of mappo in cooperative,
  multi-agent games. arXiv preprint arXiv:2103.01955  (2021)

\end{thebibliography}

\newpage
\appendix

\section{Omitted Algorithm}
\label{app:alg}
\Cref{alg:viper} shows the full pseudocode for the single-agent version of VIPER~\cite{bastani2018verifiable}.
\begin{algorithm}[tb]
\caption{VIPER for Single-Agent Setting}
\label{alg:viper}
\textbf{Input}: 
$(S, A, \transitionprob, R)$, $\pi^*$, $Q^*$, $\ntrajectories$, $\niters$%
\\
\textbf{Output}: $\hat{\pi}$ 
\begin{algorithmic}[1] 
\STATE Initialize dataset $\dataset \gets \emptyset$
\STATE Initialize policy $\hat{\pi}^0 \gets \pi^*$
\FOR{$\iter=1$ to $\niters$}
\STATE Sample $\ntrajectories$ trajectories: 

$\dataset^\iter \gets \{(s, \pi^*(s)) \sim d^{\hat{\pi}^{\iter-1}} \}$ \label{algline:vipersample}
\STATE Aggregate dataset $\dataset \gets \dataset \cup \dataset^\iter$
\STATE Resample dataset according to loss: 

$\dataset' \gets \{ (s, a) \sim p((s,a)) \propto \tilde{l}(s) \mathbb{I}[(s,a) \in \dataset] \}$ \label{algline:viperresample}
\STATE Train decision tree $\hat{\pi}^\iter \gets $ TrainDecisionTree($\dataset'$)
\ENDFOR
\STATE \textbf{return} Best policy $\hat{\pi} \in \{\hat{\pi}^1, ..., \hat{\pi}^\niters \}$ on cross validation
\end{algorithmic}
\end{algorithm}

\section{Experimental Details}

\subsection{Environments}
\label{app:env}
For all environments, we utilize the initialization and reward scheme as described in the original MADDPG paper~\cite{lowe2017multi} and Pytorch implementation.
The only change we make is to the predator-prey environment, which we describe below. 

\paragraph{Predator-prey.}
We follow the definition of the original environment proposed in the multi-agent particle environment~\cite{lowe2017multi}, with only changes in the partial observation provided to each agent. The observations of the adversary and the agents consist of the concatenation of the following vectors:
\begin{enumerate}
    \item $[\text{self}\_\text{pos}, \text{self}\_\text{vel}]$
    \item binarized relative positions and relative velocity (if applicable) of the landmarks and other agents using $sgn(x)$ as the binarizing function
    \item binarized relative distance between all pairs of agents on the other team. If the opponent team has agent $a_1, a_2$, then it will be $[sgn(x_1-x_2), sgn(y1-y2]$.
    \item binarized relative distance between all pairs of agents on the same team.
\end{enumerate}
For an environment with $K=M=2$, the observation size will be $22$ and $24$ respectively for the adversary and the agents.

\subsection{Implementation Details}
\label{app:imp}
To optimize running speed, \maviper{} adopts a caching mechanism to avoid training a new decision tree for each data point being predicted. It also does parallelization for the $\texttt{Predict}$ function starting from \Cref{alg: maviper_predict} by precomputing all the prediction information upfront, where the each prediction is delayed until all data points are looped over. In this way, \maviper{} can gather all the predictions that a particular tree needs to make and therefore do it in a parallel manner.

Since \iviper{} is fully decentralized, training of each \dt{} can be performed in parallel.

For Fitted Q-Iteration, we bin the states into 10 (mostly) evenly-spaced bins: \begin{multline*}
(-\infty, -1.), [-1., -.75), [-.75, -.5), [-.5, -.25), [-.25, 0.), \\ [0., .25),  [.25, .5), [.5, .75), [.75, 1.), (1., \infty).
\end{multline*}
We note that Fitted Q-Iteration may perform better with a better choice of bins; however, choosing the correct bin values requires either domain knowledge or extensive manual tuning to find the right balance between granularity, number of timesteps for training, and performance.

\subsection{Hyperparameters}
\label{app:hyperparams}

We vary the hyperparameters that would impact training performance of these algorithms 2-3 times and choose the hyperparameters that yield the agents with the best performance.
For all environments, we vary the number of rollouts to be $[50, 100]$ and the number of iterations to be $[30, 100]$, while the threshold is fixed at $\nagents - 1$. 
For the baselines, we also vary the maximum number of samples used for training each agent between $[10000, 30000, 100000]$.
For \imitationbaseline, we did not see much of a performance increase between $3000$ and $10000$ samples, so we pick the maximum value for fairness of comparison. 
For Fitted Q-Iteration, we also did not see much performance increase after $30000$ samples, so we chose $30000$ samples due to time constraints.
\Cref{tab:hyperparameters} shows the values of the hyperparameters that are utilized by all algorithms.
Although we set a maximum number of training iterations for \maviper, we stop training early when there is no noticeable performance gain to further improve runtime.
\begin{table}[t]
    \centering
    \begin{tabular}{ccccccc}
       Algorithm & Environment  & Max Training  & Number of & Threshold & Max Samples \\
        & & Iterations  & Rollouts \\
       \toprule 
       \iviper & Physical  & 50 & 50 \\
     & Deception & & \\
       & Cooperative  & 100 & 50 & N/A & 300,000 \\
       & Navigation && \\
       & Predator-prey & 100 & 100 \\ 
       \hline 
       \maviper & All & 100 & 50 & $\nagents - 1$ & 300,000\\
       \hline 
       \imitationbaseline & All & 20 & N/A & N/A & 100,000 \\ 
       \hline 
       Fitted Q-Iteration & All & 10 & N/A & N/A &30,000 \\ 
       \bottomrule 
    \end{tabular}
    \caption{Hyperparameter values used for all algorithms.}
    \label{tab:hyperparameters}
\end{table}




\section{Additional Results}
\label{app:results}

In \Cref{sec:indiv_perf_rew,sec:joint_perf_rew}, we further present results using the defined environment reward. 
This reward is not as intuitive as the primary metric for many of these environments, but we present the results here for the sake of completeness. 
The individual and joint performance ratios are defined in the same way as in \Cref{sec:results} in the main body of the paper, with one caveat.
Since we are now measuring reward, which may be negative or positive, we take $\frac{||A| - |B||}{A}$ to report how much more or less $B$ is than $A$.

\subsection{Individual Performance}
\label{sec:indiv_perf_rew}
\begin{figure}[t!]
    \centering 
    \begin{subfigure}[b]{1.0\textwidth}
        \centering 
        \begin{subfigure}[b]{0.42\textwidth}
        \includegraphics[width=1.0\textwidth]{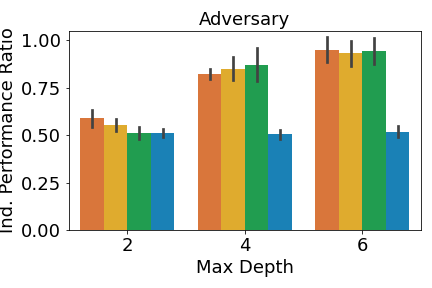}
        \end{subfigure}
        \centering
        \begin{subfigure}[b]{0.42\textwidth}
        \includegraphics[width=1.0\textwidth]{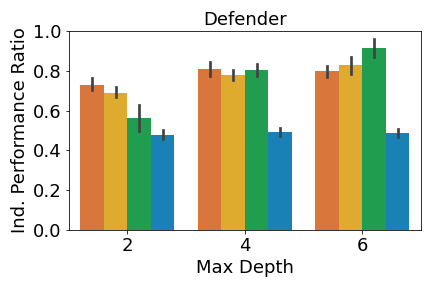}
        \end{subfigure}
        \caption{Physical Deception}
        \label{fig:pd_indiv_reward}
    \end{subfigure}
    \begin{subfigure}[b]{0.4\textwidth}
        \includegraphics[width=1.0\textwidth]{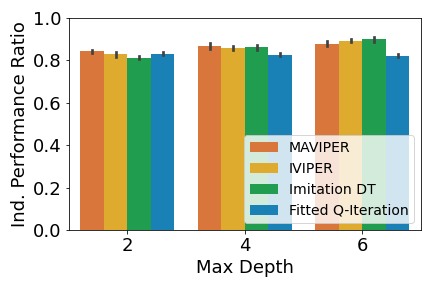}
        \caption{Cooperative Navigation}
        \label{fig:cn_indiv_reward}
    \end{subfigure}
    \begin{subfigure}[b]{1.0\textwidth}
        \centering 
        \begin{subfigure}[b]{0.42\textwidth}
        \includegraphics[width=1.0\textwidth]{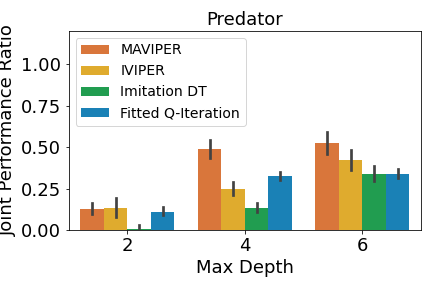}
        \end{subfigure}
        \centering
        \begin{subfigure}[b]{0.42\textwidth}
        \includegraphics[width=1.0\textwidth]{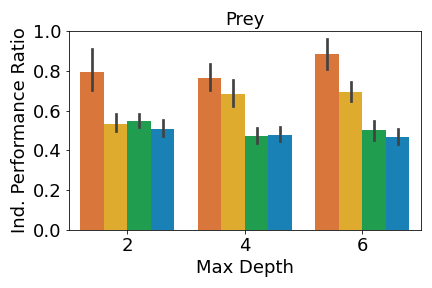}
       \end{subfigure}
        \caption{Predator-prey}
        \label{fig:pp_indiv_reward}
    \end{subfigure}
    \caption{Individual performance ratio measured by reward. Individual performance ratio of DT agents compared to expert agents for different maximum depths. Higher is better. Error bars represent the 95\% confidence interval.
    } 
    \label{fig:indiv_reward}
\end{figure}

\Cref{fig:indiv_reward} shows the individual performance ratio measured by reward on all three environments.

\paragraph{Physical Deception} 
Results for physical deception are shown in \Cref{fig:pd_indiv_reward}. 
Interestingly, \maviper{} and \iviper{} defenders only significantly outperform both baselines when the maximum depth is 2.
When the maximum depth is 4, \maviper, \iviper, and \imitationbaseline{} perform similarly, with Fitted Q-Iteration barely reaching above .50 for all depths.
When the maximum depth is 6, \imitationbaseline{} actually achieves the highest defender reward, significantly outperforming all other algorithms.
However, as shown in \Cref{fig:indiv} in the main body, \maviper{} significantly outperforms all algorithms for all maximum depths when reporting the success ratio.
This is because the reward is in part dependent on the performance of the adversary.
In other words, a poorly-performing defender can achieve similar performance to a high-performing defender if the poor-performing defender is paired with a high-performing adversary. 
We see a similar pattern for the adversary performance: \iviper, \maviper, and Fitted Q-Iteration all significantly outperform the Fitted Q-Iteration baseline for different maximum depths. 
We note that \maviper{} tends to perform better for lower maximum depths, which is desirable for interpretability.

\paragraph{Cooperative Navigation}
Results for cooperative navigation are shown in \Cref{fig:cn_indiv_reward}.
\iviper{} and \maviper{} significantly outperform the Fitted Q-Iteration baseline for all maximum depths.
However, \maviper{} only significantly outperforms \imitationbaseline{} when the maximum depth is 2. 
Otherwise, \iviper, \maviper, and \imitationbaseline{} all perform similarly. 

\paragraph{Predator-prey} 
Results for predator-prey are shown in \Cref{fig:pp_indiv_reward}.
\maviper{} prey significantly outperform all other algorithms for maximum depths of 2 and 6.
For a maximum depth of 4, \maviper{} and \iviper{} algorithms both significantly outperform the baselines.
In contrast, \maviper{} predators only significantly outperform all other algorithms for a maximum depth of 4.
For a maximum depth of 2, \maviper{} and \iviper{} significantly outperform the baselines,
For a maximum of depth of 6, \maviper, \iviper, and \imitationbaseline{} significantly outperform Fitted Q-Iteration.
Again, we note that this performance is not necessarily reflected in the results using the collision metric in \Cref{fig:pp_indiv}.

\subsection{Joint Performance}
\label{sec:joint_perf_rew}

\begin{figure}[t!]
    \centering 
    \begin{subfigure}[b]{0.4\textwidth}
        \includegraphics[width=1.0\columnwidth]{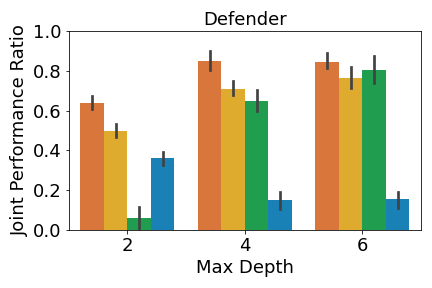}
        \caption{Physical Deception}
        \label{fig:pd_joint_reward}
    \end{subfigure}
    \begin{subfigure}[b]{0.4\textwidth}
        \includegraphics[width=1.0\textwidth]{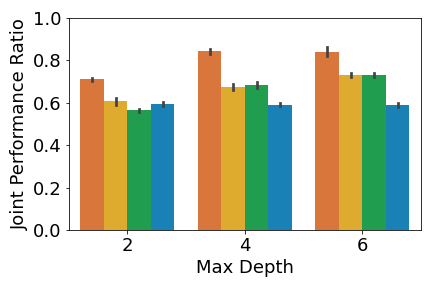}
        \caption{Cooperative Navigation}
        \label{fig:cn_joint_reward}
    \end{subfigure}
    \begin{subfigure}[b]{1.0\textwidth}
        \centering 
        \begin{subfigure}[b]{0.4\textwidth}
        \includegraphics[width=1.0\textwidth]{new/pp_joint_adv_perf_newcolors.png}
        \end{subfigure}
        \centering
        \begin{subfigure}[b]{0.4\textwidth}
        \includegraphics[width=1.0\textwidth]{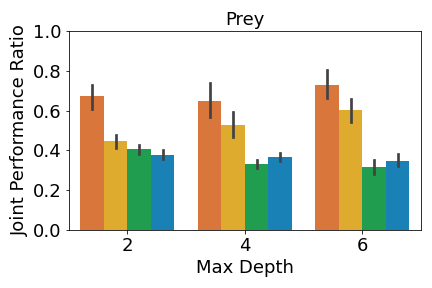}
       \end{subfigure}
        \caption{Predator-prey}
        \label{fig:pp_joint_reward}
    \end{subfigure}
    \caption{Joint performance ratio, measured by reward, of DT agents compared to expert agents for different maximum depths. DT agents are evaluated jointly. Error bars represent the 95\% confidence interval. Higher is better.
    } 
    \label{fig:joint_reward}
\end{figure}

\Cref{fig:joint_reward} shows the joint performance ratio measured by reward on all three environments.

\paragraph{Physical Deception} 
\Cref{fig:pd_joint_reward} shows the results on physical deception. 
\maviper{} defenders significantly outperform all other algorithms on this environment for maximum depths of 2 and 4.
For a maximum depth of 6, \iviper, \maviper, and \imitationbaseline all perform similarly.
Note that \maviper{} again achieves good performance for lower maximum depths, demonstrating its promise as an algorithm for producing interpretable policies.
We also note that, again, the reward metric is somewhat deceptive: when measuring the success conditions in the environment (as in \Cref{fig:joint} in the main body of the paper), \maviper{} significantly outperforms all other algorithms for all maximum depths.

\paragraph{Cooperative Navigation}
\Cref{fig:cn_joint_reward} depicts the joint agent performance on the cooperative navigation environment.
Interestingly, we see that \maviper{} significantly outperforms all other algorithms for all maximum depths, despite obtaining similar individual performance.
Consequently, this means that \maviper{} better captures the desired coordinated behavior than all of the other algorithms.

\paragraph{Predator-prey}
\Cref{fig:pp_joint_reward} shows the results for the predator-prey environment.
\maviper{} prey significantly outperform other algorithms for a maximum depth of 2.
For maximum depths of 6 and 8, it achieves slightly better (but not statistically significant) performance than \iviper, and significantly outperforms the two baselines.
\maviper{} and \iviper{} predators enjoy similar performance for maximum depths of 2 and 6.
For a maximum depth of 4, \maviper{} significantly outperforms all other algorithms.
Note that the correct behavior in this environment is challenging to capture with a small decision tree, as the number of features is either $22$ or $24$, depending on the agent type.

\subsection{Robustness to Different Opponents}
We present the full robustness results for the predator-prey and physical deception environments. 
For space reasons, we only report the average over the $100$ trials; however, we only label the best-performing agent of each type in either \textcolor{red}{red} or \textcolor{blue}{blue} if the 95\% confidence intervals do not overlap, unless otherwise mentioned. 
We exclude MADDPG from this calculation, since we know that MADDPG agents will outperform all other agent types, and we are mostly interested in how well the \textit{decision tree} policies perform. 

\begin{table}[t]
    \centering
    \begin{tabular}{ccccccc}
    \multicolumn{2}{c}{\multirow{2}{*}{}}  &   \multicolumn{5}{c}{Prey}  \\
        \cmidrule(lr){3-7}
    \multicolumn{2}{c}{}                                 &   \maviper        &   \iviper & Imitation & Fitted & MADDPG \\
      & Predator & & & DT & Q-Iteration & \\
        \hline
    \multirow{5}{*}{}   & \maviper           & (\textcolor{red}{2.28}, \textcolor{blue}{2.28}) & (\textcolor{red}{3.49}, 3.49) & (\textcolor{red}{2.41}, 2.41) & (\textcolor{red}{3.01}, 3.01) & (\textcolor{red}{1.37}, 1.37)  \\
                        & \iviper            & (1.95, \textcolor{blue}{1.95}) & (2.46, 2.46) & (2.17, 2.17) & (2.44, 2.44) & (0.88, 0.88)                 \\
                        & \imitationbaseline & (1.32, 1.32) & (1.17, \textcolor{blue}{1.17}) & (1.18, 1.18) & (1.40, 1.40) & (0.61, 0.61)                        \\
                        & Fitted Q-Iteration & (0.46, 0.46) & (0.30, 0.30) & (0.24, 0.24) & (0.18, \textcolor{blue}{0.18}) & (0.14, 0.14)                        \\      
                        & MADDPG             & (2.78, \textcolor{blue}{2.78}) & (3.36, 3.36) & (5.82, 5.82) & (4.98, 4.98) & (2.54, 2.54)                     \\        
        \bottomrule
    \end{tabular}
\caption{Robustness results of \dt{} agents on predator-prey. Results are presented as: average number of touches in an episode. Higher is better for predator, and lower is better for prey. Excluding MADDPG, the best-performing prey (lowest in value) for each predator type is in \textcolor{blue}{blue}  and the best-performing predator (highest in value) for each prey type is in \textcolor{red}{red}. }
\label{tab:robustness_pp}
\end{table}

\paragraph{Predator-prey}
\maviper{} predators are strictly more robust than all other agents (except MADDPG) to different types of prey. \maviper{} prey are the most or second most robust to different types of predators. 
In this environment, predator coordination is more critical, as predators must strategically catch the prey. The prey, on the other hand, does not require much coordination, which explains the Imitation DT prey's robustness by imitating the action of the single-agent expert.

\begin{table}[t]
\centering
\begin{tabular}{ccccccc}
\multicolumn{2}{c}{\multirow{2}{*}{}}  &   \multicolumn{5}{c}{Defender}  \\
    \cmidrule(lr){3-7}
\multicolumn{2}{c}{}                    &   \maviper    &   \iviper & Imitation & Fitted & MADDPG \\
&  Adversary & & & DT & Q-Iteration & \\
    \hline
\multirow{5}{*}{ }  & \maviper            &   (\textcolor{red}{.42}, \textcolor{blue}{.76}) & (\textcolor{red}{.45}, .33) & (.45, .23) & (\textcolor{red}{.37}, .01) & (.40, .93) \\
                    &   \iviper           &   (.39, \textcolor{blue}{.78}) & (\textcolor{red}{.45}, .32) & (.40, .23) & (\textcolor{red}{.38}, .00) & (.43, .92) \\
                    & \imitationbaseline  &   (.40, \textcolor{blue}{.79}) & (.42, .34) & (\textcolor{red}{.46}, .26) & (\textcolor{red}{.38}, .01) & (\textcolor{red}{.46}, .92) \\
                    & Fitted Q-Iteration  &   (.07, \textcolor{blue}{.77}) & (.06, .33) & (.07, .19) & (.08, .00) & (.08, .79) \\
                    & MADDPG              &   (.71, \textcolor{blue}{.76}) & (.77, .32) & (.77, .26) & (.58, .00) & (.62, .90) \\
    \bottomrule
\end{tabular}
\caption{Robustness results of \dt{} agents on physical deception. Results are presented as: (adversary success ratio, defender success ratio). Higher is better. Excluding MADDPG, the best-performing defender for each adversary type is in \textcolor{blue}{blue} and the best-performing adversary for each defender type is in \textcolor{red}{red}.}
\label{tab:robustness_pd}
\end{table}

\paragraph{Physical Deception}
\maviper{} defenders are the most robust than all agents (except MADDPG) to different types of adversaries.
Interestingly, \maviper{}, \iviper{}, and \imitationbaseline{} adversaries all perform similarly.
Indeed, they often do not achieve performance that is statistically significant from one another, as measured by the 95\% confidence interval. 
However, we still highlight the best-performing adversary in \textcolor{red}{red} to more easily show the attained performance.
Note that MADDPG adversaries can occasionally achieve success greater than around $.50$, which means that these adversaries can take advantage of some information about the defenders to correctly choose the target to visit. 
In contrast, adversaries trained with any of the \dt-learning algorithms never achieve greater than $.50$, which indicates that they may need a more complex representation to capture important details about the defenders.

\subsection{Exploitability}
\label{sec:results_exploitability} 
\begin{figure}[t!]
    \centering 
    \begin{subfigure}[b]{1.0\textwidth}
        \centering 
        \begin{subfigure}[b]{0.4\textwidth}
        \includegraphics[width=1.0\textwidth]{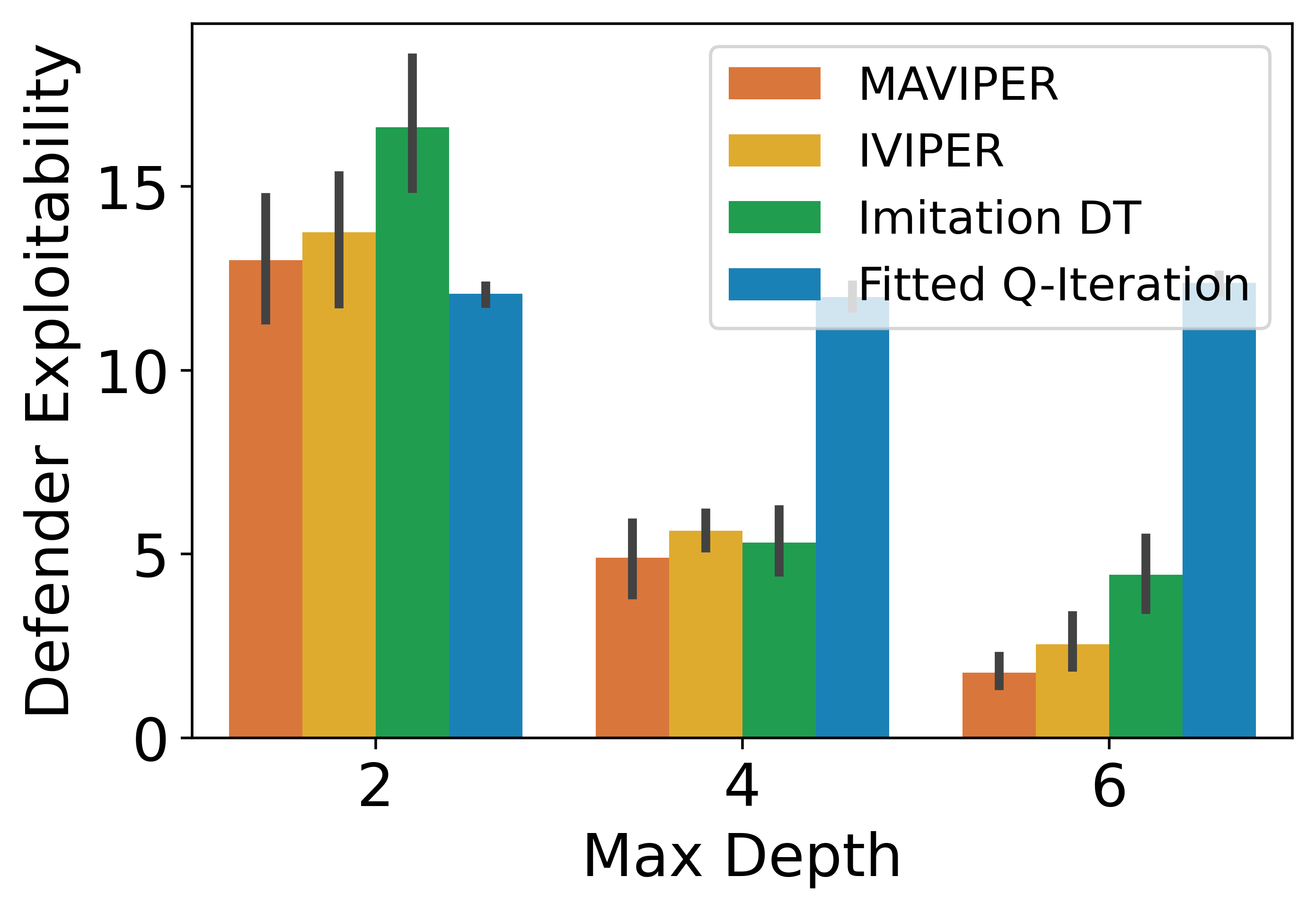}
        \end{subfigure}
        \centering
        \begin{subfigure}[b]{0.4\textwidth}
        \includegraphics[width=1.0\textwidth]{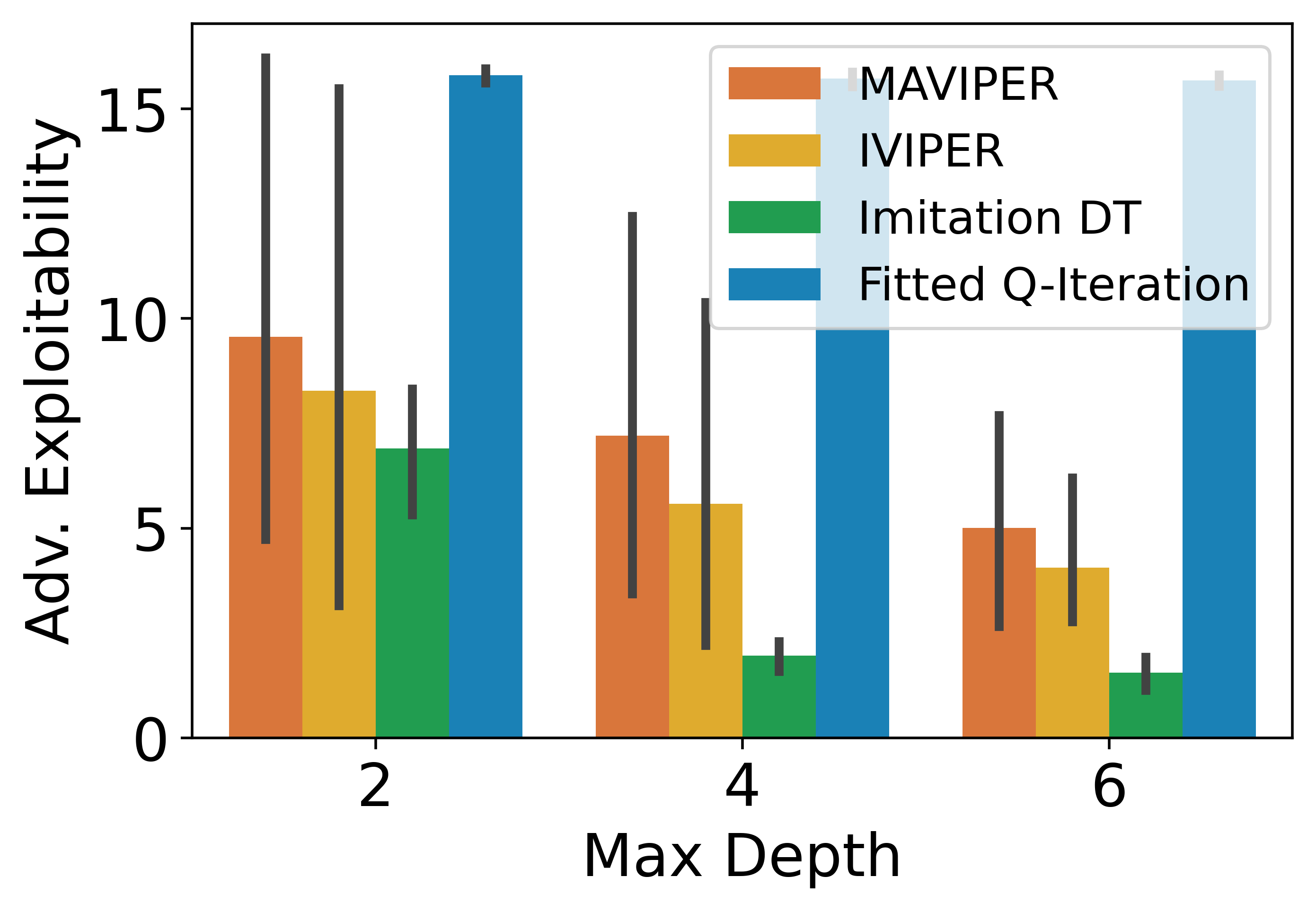}
        \end{subfigure}
    \end{subfigure}
    \caption{Exploitability of \dt{} defenders and adversaries in the physical deception environment. Lower exploitability is better.}
    \label{fig:exploitability_pd}
\end{figure}

In this set of experiments, we evaluate the exploitability of \dt~policies in the physical deception environment. Formally, we define the exploitability of a team $\team$ as:
\begin{equation}
    \text{Exploitability of team $\team$} = \max_{\pi_{-\team}'} U_{-\team}(\pi_\team, \pi_{-\team}') - U_{-\team}(\pi_\team, \pi_{-\team}),
\end{equation}
where the optimal $\pi_{-\team}'$ is the best response policy profile to team $\team$'s policies. 
Practically, to measure exploitability, we fix the policies of the agents in team $\team$ under evaluation and calculate its approximate best response $\pi_{-\team}'$ by training a new neural network policy to convergence using MADDPG.

We evaluate the exploitability of the adversary and defenders in the physical deception environment and report the restuls in \Cref{fig:exploitability_pd}. 
For the defending team, \maviper{} exhibits the lowest exploitability on all three depths, showing its effectiveness in learning coordinated policies. It is worth noting that in such a multi-agent learning setting, Imitation DT no longer performs well due to the complexity of the expert and the necessity of cooperation between agents.
For the adversary, Imitation DT performs well while \iviper{} and \maviper{} performs similarly as the second best. This could be the result that imitation learning quickly imitates a near-optimal expert adversary starting from the depth of four. Since the adversary consists of only a single agent, \maviper{} and \iviper{} are reduced to the same method. 

\end{document}